\pdfoutput=1
\documentclass[11pt]{article}

\usepackage[]{acl}

\usepackage{times}
\usepackage{latexsym}

\usepackage[T1]{fontenc}
\usepackage[utf8]{inputenc}

\usepackage{microtype}

\usepackage{inconsolata}

\title{Entity-level Factual Adaptiveness of Fine-tuning based Abstractive Summarization Models}

\author{Jongyoon Song$^{1}$   Nohil Park$^{1}$    Bongkyu Hwang$^{2}$     Jaewoong Yun$^{2}$ \\ {\bf Seongho Joe$^{2}$} {\bf Youngjune L. Gwon$^{2}$} {\bf Sungroh Yoon$^{1, 3,4}$\Thanks{\hspace{0.1em} Corresponding author}} \\
   $^{1}$Data Science \& AI Laboratory, Seoul National University, Korea \\
   $^{2}$Samsung SDS, Korea \\
   $^{3}$Deptment of ECE and Interdisciplinary Program in AI, Seoul National University, Korea \\
   $^{4}$ASRI, INMC, and AIIS, Seoul National University, Korea \\
   \{\texttt{coms1580}, \texttt{pnoil2588}, \texttt{sryoon}\}\texttt{@snu.ac.kr} \\
   \{\texttt{bongkyu.hwang}, \texttt{jw0531.yun}, \texttt{drizzle.cho}, \texttt{gyj.gwon}\}\texttt{@samsung.com} \\ }

\begin{document}
\maketitle
\begin{abstract}
Abstractive summarization models often generate factually inconsistent content particularly when the parametric knowledge of the model conflicts with the knowledge in the input document.
In this paper, we analyze the robustness of fine-tuning based summarization models to the knowledge conflict, which we call \textit{factual adaptiveness}.
We utilize pre-trained language models to construct evaluation sets and find that factual adaptiveness is not strongly correlated with factual consistency on original datasets.
Furthermore, we introduce a controllable counterfactual data augmentation method where the degree of knowledge conflict within the augmented data can be adjustable.
Our experimental results on two pre-trained language models (PEGASUS and BART) and two fine-tuning datasets (XSum and CNN/DailyMail) demonstrate that our method enhances factual adaptiveness while achieving factual consistency on original datasets on par with the contrastive learning baseline.
\end{abstract}

\section{Introduction}\label{sec:introduction}
Factual consistency is a crucial aspect, especially in abstractive summarization, ensuring that the facts presented in the generated summary align with those in the input document (\citealp{maynez-etal-2020-faithfulness}; \citealp{kryscinski-etal-2020-evaluating}; \citealp{huang-2021-factual}; \citealp{scialom-etal-2021-questeval}; \citealp{fabbri-etal-2022-qafacteval}). %

Recent summarization models using pre-training and/or fine-tuning of the language model have shown excellent performance in various aspects such as factual consistency (\citealp{lewis-etal-2020-bart}; \citealp{raffel-2020-exploring}; \citealp{zhang-2020-pegasus}; \citealp{cao-wang-2021-cliff}; \citealp{wan-bansal-2022-factpegasus}; \citealp{roit-2023-factually}).
There are also studies on large language models (\citealp{brown-2020-language}; \citealp{ouyang-2022-training}; \citealp{chowdhery-2022-palm}) for the summarization (\citealp{zhang-2023-extractive}; \citealp{adams-2023-sparse}) or the evaluation of summaries (\citealp{luo-2023-chatgpt}; \citealp{gao-2023-human}).

Previous works have also reported that (large) language models have \textit{parametric knowledge} (\citealp{ji-2023-survey}; \citealp{bang-2023-multitask}).
The parametric knowledge of the language model is known to result in hallucinated contents, particularly when \textit{knowledge conflict} occurs which refers to the mismatch between the knowledge in the document and the parametric knowledge of the model (\citealp{longpre-etal-2021-entity}; \citealp{neeman-2022-disentqa}; \citealp{zhou-2023-context}).
Because the hallucination problem degrades factual consistency of summarization models (\citealp{maynez-etal-2020-faithfulness}; \citealp{nan-etal-2021-entity}), it is important to study the robustness to knowledge conflict of summarization models.

In abstractive summarization, most previous works have measured the factual consistency using the original document only, which is not sufficient to evaluate the robustness to knowledge conflict (\citealp{cao-wang-2021-cliff}; \citealp{wan-bansal-2022-factpegasus}; \citealp{wan-etal-2023-faithfulness}).
There are studies on hallucination problems caused by knowledge conflict in abstractive summarization (\citealp{ladhak-etal-2023-pre}; \citealp{cheang-2023-temposum}). 
However, the aforementioned document perturbation strategies do not control the degree of knowledge conflict, which offers valuable insight into the robustness of the summarization models to the knowledge conflict.

In this paper, we define \textit{factual adaptiveness}, the robustness to the knowledge conflict, of fine-tuning based abstractive summarization models.
We focus on entity-level knowledge conflict and factual adaptiveness and use counterfactual samples obtained by replacing a single named entity (i.e., original entity) with another named entity (i.e., counterfactual entity).

Unlike previous works on knowledge conflict in question answering, there are two additional considerations in our work (\citealp{longpre-etal-2021-entity}; \citealp{neeman-2022-disentqa}).
First, we determine which named entity to replace in the reference summary by detecting parametric knowledge.
Second, we select the named entity to be replaced with to control knowledge conflict.
To address those considerations, we utilize the parametric knowledge of the pre-trained language model (PLM) during the knowledge conflict set construction.

We first analyze the factual adaptiveness of various methods for improving factual consistency on original datasets such as data filtering (\citealp{nan-etal-2021-entity}), contrastive learning (\citealp{cao-wang-2021-cliff}), and advanced decoding (\citealp{wan-etal-2023-faithfulness}).
Our results demonstrate that methods for factual consistency on original datasets do not always effectively mitigate knowledge conflict problems, which indicates that factual consistency on original datasets can be orthogonal to factual adaptiveness.

We next propose a controllable counterfactual data augmentation technique.
Specifically, the method constructs counterfactual samples based on a pre-defined degree of knowledge conflict.
Experimental results show that our method improves factual adaptiveness effectively and addresses the entity-level hallucination problem caused by knowledge conflict.

Our contributions can be summarized as follows:
\begin{itemize}
    \item We introduce the factual adaptiveness of fine-tuning based summarization models using a parametric knowledge of a pre-trained language model.
    \item We demonstrate that factual consistency on original datasets tends to be orthogonal to factual adaptiveness.
    Specifically, data filtering largely improves factual adaptiveness while advanced decoding and contrastive learning show minimal differences.
    \item We propose a controllable counterfactual data augmentation method that enhances factual adaptiveness while preserving factual consistency on original datasets.
\end{itemize}

\begin{figure*}[!t]
  \includegraphics[width=0.95\textwidth]{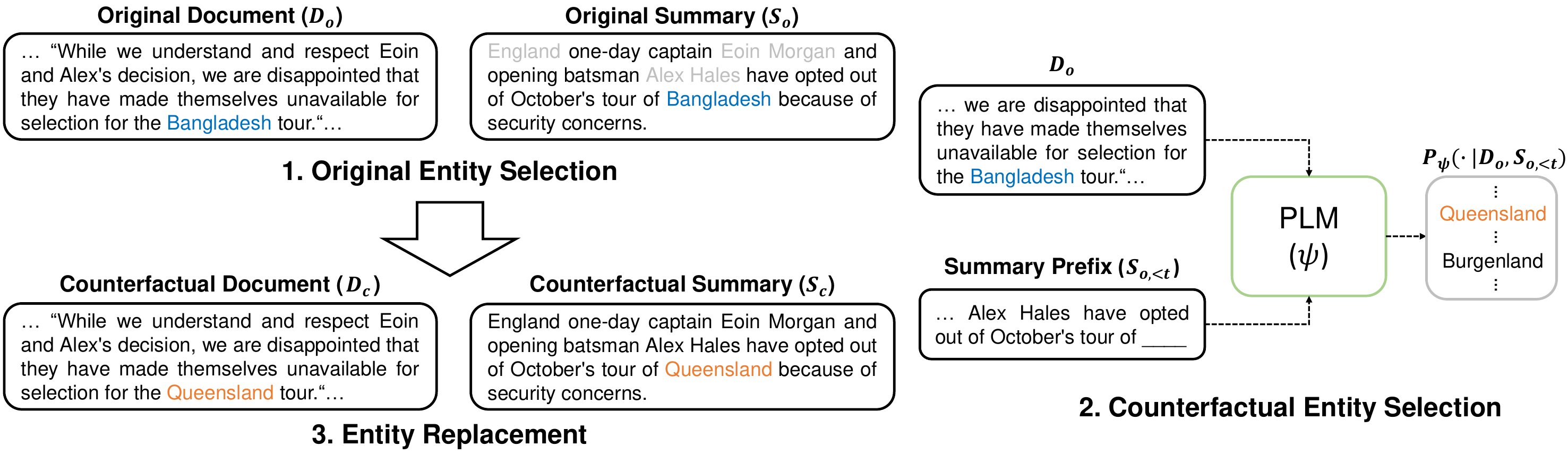}
  \caption{Overview of the counterfactual sample construction process.
  The example is sampled from the XSum validation set.}
  \label{fig:overview}
\end{figure*}

\section{Factual Adaptiveness}\label{sec:factual_adaptiveness}
In this section, we define and analyze factual adaptiveness of various fine-tuning based summarization models which are known to improve factual consistency.
We formulate factual adaptiveness in Section \ref{subsec:factual_adaptiveness:formulation}, and explain the factual adaptiveness evaluation set construction method in Section \ref{subsec:factual_adaptiveness:evaluation_set_construction}.
In the remaining text, the term \textit{counterfactual} indicates the presence of knowledge conflict caused by the entity replacement.
We also denote a \textit{counterfactual sample} as a pair of the counterfactual document and summary, assuming they are factually consistent.

\subsection{Formulation}\label{subsec:factual_adaptiveness:formulation}
Suppose we have a sample $X_o=(D_o, S_o)$ which consists of document $D_o=\{d_1, d_2, ..., d_M\}$ and a reference summary $S_o=\{s_1, s_2, ..., s_T\}$.
We denote a pre-trained language model as $\psi$ and a fine-tuned summarization model as $\phi$.

To construct a counterfactual sample $X_c=(D_c, S_c)$ from $X_o$, we first select the \textbf{original named entity} $E_o$ which (\textit{i}) exists in both $D_o$ and $S_o$ and (\textit{ii}) contains the parametric knowledge of $\psi$.
We then replace $E_o$ with the \textbf{counterfactual named entity} $E_c$ to synthesize $X_c$ which consists of the counterfactual document $D_c$ and factually consistent summary $S_c$.

We define factual adaptiveness metrics $M_{CL}$ and $M_{FC}$ on two perspectives: \textbf{conditional likelihood} and \textbf{factual consistency}, respectively.
Specifically, we input original and counterfactual documents alternately into the summarization model, measuring two distinct differences: i) the conditional likelihood of original (counterfactual) named entities within the reference summary and ii) the factual consistency between the original (counterfactual) document and the generated summary.

We define $M_{CL}$ as follows:
\begin{equation}\label{eq:m_cl}
    M_{CL} := P_{\phi}(e_o|D_o,S_{o,<t}) - P_{\phi}(e_c|D_c,S_{c,<t}),
\end{equation}
where $S_{c,<t}$ and $S_{o,<t}$ denote the summary prefix of first $t-1$ tokens of $S_c$ and $S_o$, respectively. 
$e_c$ and $e_o$ denote the first tokens of $E_c$ and $E_o$, respectively, assuming that $e_c$ and $e_o$ are $t$-th tokens of each summary. 
$M_{CL}$ indicates the factual adaptiveness of model $\phi$ on the perspective of the conditional likelihood when the counterfactual document and the summary prefix are given.

Because $M_{CL}$ does not consider the summary generated by $\phi$, we introduce complementary metric $M_{FC}$ as follows:
\begin{equation}\label{eq:m_fc}
    M_{FC} := f(D_o, S^{\phi}(D_o)) - f(D_c, S^{\phi}(D_c)),
\end{equation}
where $f$ denotes factual consistency scoring function such as QuestEval (\citealp{scialom-etal-2021-questeval}), and $S^{\phi}(D)$ denotes the summary generated by $\phi$ given the document $D$.
The second term of $M_{FC}$ involves inputting documents where a knowledge conflict occurs, leading to the generation of factually inconsistent summaries from the model. 
As a result, $M_{FC}$ approximates factual adaptiveness by calculating the reduction in the factual consistency of the summarization model due to knowledge conflicts.

In the remaining text, we refer to \textit{factual consistency} as the attribute between the \textbf{original} document and the generated summary if further clarification is not provided.
Note that factual consistency is different from $M_{FC}$ which measures the \textit{difference} of factual consistency scores between original and counterfactual samples.

\begin{algorithm}[t!]
\caption{Entity Validation Scenario (S1)}\label{alg:entity_selection_scenario_1}
\textbf{Input: } {Document $D_o=\{d_1, d_2, ..., d_M\}$, summary $S_o=\{s_1, s_2, ..., s_T\}$, pre-trained language model $\psi$, null document $D_\varnothing$, threshold $\tau$.} \\
\textbf{Output: } {Counterfactual samples $X_c$} \\
\begin{algorithmic}[1]
\newcommand*{\Break}{\textbf{break}}
\State $X_c = \{\}$
\State $E = \{\}$
\State {Get $L=\{E_{o,1}, E_{o,2}, ..., E_{o,K}\}$, the list of named entity which exists in both $D_o$ and $S_o$}
\For{\texttt{$k \gets 1$ to $K$}}
    \State $t_k \gets $ {the first token position of $E_{o,k}$ in $S_o$}
    \State $E_c \gets $ {named entity sampled from one of three groups}\Comment{Section \ref{subsubsec:factual_adaptiveness:evaluation_set_construction:counterfactual_entity_candidates}}
    \State $p \gets P_{\psi}(s_{t_k}|D_\varnothing, S_{o,<{t_k}})$
    \If {$p > \tau$}\Comment{Section \ref{subsubsec:factual_adaptiveness:factual_adaptiveness:evaluation_set_construction:original_and_counterfactual_named_entity}}
        \State {Append $(E_{o,k}, E_c)$ to $E$}
        % \State \Break
    \EndIf
\EndFor
\For{each pair $(E_o, E_c)$ in $E$}\Comment{Section \ref{subsubsec:factual_adaptiveness:factual_adaptiveness:entity_replacement}}
    \State $D_c\gets$ \Call{Replace}{$D_o,E_o,E_c$}
    \State $S_c\gets$ \Call{Replace}{$S_o,E_o,E_c$}
    \State {Append $(D_c, S_c)$ to $X_c$}
\EndFor
\State \Return $X_c$
\end{algorithmic}
\end{algorithm}

\subsection{Evaluation Set Construction}\label{subsec:factual_adaptiveness:evaluation_set_construction}
To satisfy our assumption: (\textit{i}) $E_o$ contains the parametric knowledge of the model and (\textit{ii}) $D_c$ occurs knowledge conflict, it is critical to select appropriate $E_o$ and $E_c$.
We utilize PLM $\psi$ during the entity selection to accurately construct counterfactual sample $X_c$.

\subsubsection{Counterfactual Entity Candidate Pool}
We restrict the candidate entities to those of the same category and found in the training corpus following previous works (\citealp{longpre-etal-2021-entity}; \citealp{rajagopal-2022-counterfactual}).
We utilize \textit{spaCy} (\citealp{Honnibal_spaCy_Industrial-strength_Natural_2020}) to construct a candidate pool of counterfactual entities from the named entities in the fine-tuning set.

\subsubsection{Original Entity Candidates}
For each reference summary $S_o$, we extract the named entity list $L=\{E_{o,1}, E_{o,2}, ..., E_{o,K}\}$ (if $i < j$, $E_{o,i}$ appears before $E_{o,j}$ in $S_o$) using \textit{spaCy}.
In this work, we exclude numerical categories such as \texttt{QUANTITY}, \texttt{DATE}, and \texttt{TIME} concerning that numerical entities can easily be paraphrased (e.g. \texttt{15:00} / \texttt{3:00 PM}, \texttt{1970s} / \texttt{70's}).

For each named entity $E_{o,k}$, we validate that the entity is part of the parametric knowledge of $\psi$.
We hypothesize two validation scenarios which will be described in Section \ref{subsubsec:factual_adaptiveness:factual_adaptiveness:evaluation_set_construction:original_and_counterfactual_named_entity}.

\subsubsection{Counterfactual Entity Candidates}\label{subsubsec:factual_adaptiveness:evaluation_set_construction:counterfactual_entity_candidates}
We assume that the original named entity $E_{o,k}$ appears in $S_o$ at the position $t_k$.
We sort counterfactual entity candidates by the conditional likelihood of their first token given the document $D_o$ and the prefix of the reference summary $S_{o,<t_k}$.

We divide the counterfactual entity candidates into three groups: \textbf{Top} (top 2\%-25\% entities by the conditional likelihood), \textbf{Middle (Mid)} (25\%-75\%), and \textbf{Bottom (Bot)} (75\%-100\%).
Note that we exclude the top 2\% entities to ensure counterfactual replacement.
Intuitively, the degree of knowledge conflict is expected to be larger in \textbf{Bot} compared to \textbf{Top}.
We select the group and sample counterfactual entity candidate from the group before the validation step.

\subsubsection{Original and Counterfactual Entity Validation}\label{subsubsec:factual_adaptiveness:factual_adaptiveness:evaluation_set_construction:original_and_counterfactual_named_entity}
We set two scenarios for the entity validation to satisfy the assumptions described in Section \ref{subsec:factual_adaptiveness:evaluation_set_construction}.

\textbf{Scenario 1 (S1): Unconditional Likelihood}
We hypothesize that the named entity $E_{o,k}$ whose unconditional likelihood $P_\psi(e_{o,k}|D_{\varnothing}, S_{o,<{t_k}})$ surpasses the threshold $\tau$ is part of the parametric knowledge of $\psi$.
$e_{o,k}$ denotes the first token of $E_{o,k}$ (i.e., $s_{t_k}$), and $D_\varnothing$ denotes the \textit{null document} such as ".".
Note that after $E_o$ is validated, we do not further examine $E_c$ in Scenario 1.
We refer to Algorithm \ref{alg:entity_selection_scenario_1} for details.
 
\textbf{Scenario 2 (S2): Conditional Likelihood Difference}
We hypothesize that $E_{o,k}$ and $E_c$ contain parametric knowledge and knowledge conflict, respectively, if the conditional likelihood difference $P_\psi(e_{o,k}|D_{o}, S_{o,<{t_k}}) - P_\psi(e_{c}|D_{c}, S_{c,<{t_k}})$  surpasses the threshold $\tau$.
Note that the condition in Scenario 2 is directly aligned to $M_{CL}$ in Equation \ref{eq:m_cl} except for the model to be used.
The algorithm of Scenario 2 can be found in Appendix \ref{sec:appendix:algorithm2}.

\subsubsection{Entity Replacement}\label{subsubsec:factual_adaptiveness:factual_adaptiveness:entity_replacement}
If the sample $X_o$ has the valid original (counterfactual) entity $E_o$ ($E_c$), we replace all $E_o$ in $D_o$ and $S_o$ with $E_c$.
After the entity-level replacement, we further conduct the word-level replacement where each word in $E_o$ is replaced with the word in $E_c$ proportionally to its position. 

For example, if $E_o=$ \texttt{"Daniel Radcliffe"} and $E_c=$ \texttt{"Rupert Grint"}, we further replace \texttt{"Daniel"} with \texttt{"Rupert"} and \texttt{"Radcliffe"} with \texttt{"Grint"}.
\section{Analysis on Models for Improving Factual Consistency}\label{sec:analysis_on_models_for_factual_consistency}
In this section, we analyze summarization models using the evaluation set as described in Section \ref{subsec:factual_adaptiveness:evaluation_set_construction}.
Specifically, we measure $M_{CL}$ and $M_{FC}$ of various models that are proposed to improve factual consistency to observe the relation between factual adaptiveness and factual consistency.

\begin{table*}
\centering
\resizebox{\linewidth}{!}{
\begin{tabular}{l|c|c|ccc|ccc|ccc|ccc}
\hline
\toprule
\multicolumn{1}{c}{} &
\multicolumn{1}{c}{} &
\multicolumn{1}{c}{} &
\multicolumn{3}{c}{$M_{CL}$(S1)($\downarrow$)} &
\multicolumn{3}{c}{$M_{CL}$(S2)($\downarrow$)} &
\multicolumn{3}{c}{$M_{FC}$(S1)($\downarrow$)} &
\multicolumn{3}{c}{$M_{FC}$(S2)($\downarrow$)} \\
Models & R-L & QEval & Top & Mid & Bot & Top & Mid & Bot & Top & Mid & Bot & Top & Mid & Bot \\
\midrule
\multicolumn{15}{c}{XSum (PEGASUS)} \\
\hline
NLL & \textbf{36.36} & 32.94 & 0.552 & 0.589 & 0.631 & 0.718 & 0.734 & 0.744 & 1.79 & 2.10 & 1.98 & 2.14 & 2.43 & 2.55\\
& $\pm0.07$ & $\pm0.05$ & $\pm.004$ & $\pm.002$ & $\pm.006$ & $\pm.004$ & $\pm.003$ & $\pm.004$ & $\pm0.08$ & $\pm0.08$ & $\pm0.09$ & $\pm0.10$ & $\pm0.02$ & $\pm0.11$\\
Filtering & 34.89 & 33.49 & \textbf{0.495} & \textbf{0.526} & \textbf{0.565} & \textbf{0.669} & \textbf{0.681} & \textbf{0.693} & \textbf{1.59} & \textbf{1.77} & \textbf{1.75} & \textbf{1.69} & \textbf{1.91} & \textbf{2.07}\\
& $\pm0.09$ & $\pm0.24$ & $\pm.011$ & $\pm.015$ & $\pm.019$ & $\pm.010$ & $\pm.009$ & $\pm.013$ & $\pm0.00$ & $\pm0.12$ & $\pm0.12$ & $\pm0.17$ & $\pm0.08$ & $\pm0.15$\\
Decoding & 35.29 & \textbf{34.11} & - & - & - & - & - & - & 1.76 & 1.90 & 2.01 & 2.09 & 2.40 & 2.46\\
& $\pm0.02$ & $\pm0.02$ & - & - & - & - & - & - & $\pm0.11$ & $\pm0.02$ & $\pm0.14$ & $\pm0.19$ & $\pm0.09$ & $\pm0.10$\\
CLIFF & 35.86 & 33.27 & 0.547 & 0.583 & 0.625 & 0.713 & 0.727 & 0.740 & 1.83 & 2.12 & 2.11 & 2.10 & 2.30 & 2.50\\
& $\pm0.04$ & $\pm0.02$ & $\pm.005$ & $\pm.004$ & $\pm.003$ & $\pm.006$ & $\pm.005$ & $\pm.005$ & $\pm0.09$ & $\pm0.16$ & $\pm0.04$ & $\pm0.04$ & $\pm0.16$ & $\pm0.16$\\
\hline
\multicolumn{15}{c}{CNN/DailyMail (PEGASUS)} \\
\hline
NLL & 37.08 & 51.44 & 0.243 & 0.277 & 0.304 & 0.444 & 0.451 & 0.449 & 0.49 & 0.46 & 0.45 & 0.53 & 0.43 & 0.44\\
& $\pm0.05$ & $\pm0.05$ & $\pm.003$ & $\pm.001$ & $\pm.001$ & $\pm.002$ & $\pm.001$ & $\pm.002$ & $\pm0.14$ & $\pm0.04$ & $\pm0.07$ & $\pm0.19$ & $\pm0.14$ & $\pm0.07$\\
Filtering & 36.69 & 51.86 & \textbf{0.188} & \textbf{0.215} & \textbf{0.243} & \textbf{0.384} & \textbf{0.390} & \textbf{0.396} & \textbf{0.31} & \textbf{0.19} & \textbf{0.29} & \textbf{0.37} & 0.46 & 0.34\\
& $\pm0.10$ & $\pm0.03$ & $\pm.002$ & $\pm.001$ & $\pm.002$ & $\pm.003$ & $\pm.001$ & $\pm0.002$ & $\pm0.07$ & $\pm0.09$ & $\pm0.11$ & $\pm0.01$ & $\pm0.05$ & $\pm0.06$\\
Decoding & \textbf{37.52} & \textbf{52.60} & - & - & - & - & - & - & 0.54 & 0.41 & 0.48 & 0.53 & \textbf{0.31} & 0.49\\
& $\pm0.10$ & $\pm0.05$ & - & - & - & - & - & - & $\pm0.16$ & $\pm0.18$ & $\pm0.08$ & $\pm0.09$ & $\pm0.13$ & $\pm0.08$\\
CLIFF & 37.06 & 51.45 & 0.243 & 0.278 & 0.302 & 0.445 & 0.452 & 0.450 & 0.56 & 0.60 & 0.62 & 0.40 & 0.50 & \textbf{0.33}\\
& $\pm0.04$ & $\pm0.03$ & $\pm.002$ & $\pm.003$ & $\pm.003$ & $\pm.001$ & $\pm.000$ & $\pm.002$ & $\pm0.15$ & $\pm0.11$ & $\pm0.19$ & $\pm0.02$ & $\pm0.10$ & $\pm0.12$\\
\hline
\end{tabular}}
\caption{
Mean and standard deviation of ROUGE-L (R-L) and QuestEval (QEval) on original test sets and $M_{CL}$/$M_{FC}$ scores on factual adaptiveness evaluation sets across 3 seeds.
}\label{table:baseline_analysis_pegasus}
\end{table*}

\subsection{Setup}
We measure ROUGE-L (\citealp{lin-2004-rouge}) and QuestEval score, which is known to be aligned with human judgments (\citealp{scialom-etal-2021-questeval}), on the original test set and $M_{CL}$/$M_{FC}$ scores on the factual adaptiveness evaluation set.
We use three approaches for the baseline: data filtering (Filtering, \citealp{nan-etal-2021-entity}), contrastive learning (CLIFF, \citealp{cao-wang-2021-cliff}), and advanced decoding (Decoding, \citealp{wan-etal-2023-faithfulness}) and two backbone PLMs: PEGASUS\textsubscript{LARGE} (\citealp{zhang-2020-pegasus}) and BART\textsubscript{LARGE} (\citealp{lewis-etal-2020-bart}) for the analysis.
We also evaluate models that are simply fine-tuned with negative log-likelihood objectives (NLL) for comparison.
For the baseline re-implementation, we use \textit{HuggingFace}\footnote[1]{\url{https://github.com/huggingface/transformers}} for PEGASUS based models and \textit{fairseq}\footnote[2]{\url{https://github.com/facebookresearch/fairseq}} for BART based models.
Hyperparameters for each baseline can be found in Appendix \ref{sec:appendix:hyperparameters_and_dataset_stats}.

\subsection{Evaluation Set}
We use test sets of XSum (\citealp{narayan-etal-2018-dont}) and CNN/DailyMail (CNNDM, \citealp{Hermann-2015-teaching}) to construct factual adaptiveness evaluation sets.
We search the threshold $\tau$ using validation sets so that the extracted factual adaptiveness evaluation set is about 10\% of the original validation set (We use \textbf{Top} group and Scenario 1).
$\tau$ and dataset statistics can be found in Appendix \ref{sec:appendix:hyperparameters_and_dataset_stats}.

To specify the evaluation set, information on (i) the type of PLM, (ii) the dataset, (iii) the type of counterfactual entity candidate group, and (iv) the type of validation scenario is required.
For example, we denote XSum (PEGASUS, Top, S1) as the evaluation set based on the XSum test set using PEGASUS for the PLM, \textbf{Top} group for the counterfactual entity candidate group, and scenario 1 for the entity validation.

\subsection{Results}\label{subsec:analysis:results}
Scores of PEGASUS based models are shown in Table \ref{table:baseline_analysis_pegasus}.
Results on BART based models can be found in Appendix \ref{sec:appendix:baseline-analysis-bart}, except for Decoding because the original training code for BART is implemented on \textit{fairseq}, while the code for Decoding is based on \textit{HuggingFace}.

\textbf{Entity Validation Scenarios} We first observe which of the two entity validation scenarios more effectively generates knowledge conflict.
In most cases, it is observed that factual adaptiveness is much degraded for the evaluation sets constructed based on Scenario 2, especially in XSum.
The results suggest that through Scenario 2, we can accurately detect prior knowledge of PLM and effectively induce knowledge conflicts compared to Scenario 1.
Given the fact that the criterion used in Scenario 2 is similar to Equation \ref{eq:m_fc}, and they only differ in terms of the models used, we speculate that fine-tuned models share the knowledge with pre-trained models.

\textbf{Counterfactual Entity Candidate Groups} 
We can observe that $M_{CL}$ scores tend to increase in the order of \textbf{Top}, \textbf{Mid}, and \textbf{Bot}.
Considering that the group is divided based on the conditional likelihood of PLM, the results indicate that our method controls the degree of parametric knowledge and knowledge conflict effectively.

In CNN/DailyMail, the tendency for $M_{FC}$ between the candidate groups is weak compared to XSum even with the consistency in $M_{CL}$.
We speculate that the low abstractiveness of CNN/DailyMail (\citealp{dreyer-etal-2023-evaluating}) has improved overall factual adaptiveness with respect to $M_{CL}$ and $M_{FC}$, resulting in the similarity of $M_{FC}$ between the candidate groups.

\textbf{Factual Adaptiveness vs. Factual Consistency} While Filtering greatly enhances both factual consistency and factual adaptiveness, Decoding and CLIFF show minimal improvements in $M_{CL}$ and $M_{FC}$ scores compared to NLL.
The results imply that methods for factual consistency improvement do not necessarily increase robustness against knowledge conflict, and factual consistency is not strongly correlated with factual adaptiveness.

\begin{table*}
\centering
\resizebox{\linewidth}{!}{
\begin{tabular}{c|c|c|cc|c|c|cc|c|c|cc|c|c|cc}
\hline
\toprule
\multicolumn{1}{c|}{} &
\multicolumn{8}{c|}{XSum} &
\multicolumn{8}{c}{CNN/DailyMail}\\
\hline
\multicolumn{1}{c|}{} &
\multicolumn{4}{c|}{PEGASUS} &
\multicolumn{4}{c|}{BART} &
\multicolumn{4}{c|}{PEGASUS} &
\multicolumn{4}{c}{BART}\\
\hline
 & R-L & QEval & $M_{CL}$ & $M_{FC}$ & R-L & QEval & $M_{CL}$ & $M_{FC}$ & R-L & QEval & $M_{CL}$ & $M_{FC}$ & R-L & QEval & $M_{CL}$ & $M_{FC}$\\
 \hline
NLL & \textbf{36.36} & 32.94 & 0.734 & 2.43 & \textbf{34.83} & 32.94 & 0.752 & 2.14 & \textbf{37.08} & 51.44 & 0.451 & 0.43 & \textbf{38.05} & 50.99 & 0.438 & 0.62 \\
  & $\pm0.07$ & $\pm0.05$ & $\pm.003$ & $\pm0.02$ & $\pm0.05$ & $\pm0.03$ & $\pm.004$ & $\pm0.02$ & $\pm0.05$ & $\pm0.05$ & $\pm.001$ & $\pm0.14$ & $\pm0.04$ & $\pm0.04$ & $\pm.006$ & $\pm0.10$\\
  % \hline
CLIFF & 35.86 & \textbf{33.27} & 0.727 & 2.30 & 33.89 & 33.32 & 0.742 & 2.19 & 37.06 & \textbf{51.45} & 0.452 & 0.50 & 37.97 & \textbf{51.07} & 0.435 & \textbf{0.52} \\
  & $\pm0.04$ & $\pm0.02$ & $\pm.005$ & $\pm0.16$ & $\pm0.14$ & $\pm0.07$ & $\pm.005$ & $\pm0.07$ & $\pm0.04$ & $\pm0.03$ & $\pm.000$ & $\pm0.10$ & $\pm0.13$ & $\pm0.06$ & $\pm.003$ & $\pm0.10$\\
  % \hline
Ours & 35.69 & 33.26 & \textbf{0.132} & \textbf{1.20} & 33.81 & \textbf{33.39} & \textbf{0.113} & \textbf{1.20} & 36.91 & 51.37 & \textbf{0.096} & \textbf{0.41} & 37.88 & 51.01 & \textbf{0.074} & 0.56 \\
(CLIFF)  & $\pm0.03$ & $\pm0.06$ & $\pm.004$ & $\pm0.10$ & $\pm0.04$ & $\pm0.05$ & $\pm.005$ & $\pm0.10$ & $\pm0.00$ & $\pm0.04$ & $\pm.001$ & $\pm0.14$ & $\pm0.09$ & $\pm0.03$ & $\pm.001$ & $\pm0.21$\\
  \hline
\end{tabular}}
\caption{
ROUGE-L (R-L) and QuestEval (QEval) on original test sets and $M_{CL}$/$M_{FC}$ scores on factual adaptiveness evaluation sets of Scenario 2 and \textbf{Mid} group with the mean and standard deviation across 3 seeds. 
}\label{table:ours_main_results}
\end{table*}

\section{Controllable Counterfactual Data Augmentation}\label{sec:method}

\subsection{Training Set Construction}
We apply the same procedure used for building a factual adaptiveness evaluation set to construct the augmentation set.
For each dataset, we use the same threshold $\tau$ determined during the corresponding evaluation set construction.
We further proceed to sample the obtained augmentation set at a certain ratio $\rho$ of the original training set.

\subsection{Incorporation with Contrastive Learning}\label{subsec:method:training_set_construction}
In recent research, contrastive learning has been applied to enhance factual consistency (\citealp{cao-wang-2021-cliff}; \citealp{wan-bansal-2022-factpegasus}). 
Our method can be integrated with a contrastive learning-based approach if it can map positive/negative summaries to counterfactual documents.

In the context of contrastive learning, we apply previous contrastive learning set construction methods to the counterfactual samples.
For CLIFF, we utilize the provided positive/negative summaries by replacing original entities in the summaries with counterfactual entities.
For FactPEGASUS (\citealp{wan-bansal-2022-factpegasus}), we feed augmented datasets to the provided contrastive learning pipelines\footnote[3]{\url{https://github.com/meetdavidwan/factpegasus}}.

In the remaining text, the term \textbf{ours} refers to a model that integrates controllable counterfactual data augmentation with the CLIFF training method.
We also conduct experiments on FactPEGASUS and experimental results on XSum can be found in Appendix \ref{sec:appendix:factpegasus}.

\section{Experiments}\label{sec:experiments}
\subsection{Setup}\label{subsec:experiments:setup}

We use Scenario 2 and \textbf{Mid} group to construct augmented contrastive learning training sets in accordance with the conclusions drawn in Section \ref{subsec:analysis:results}.
To regulate the size of the training dataset, we sample the augmentation set from counterfactual samples, setting $\rho$ to 0.1.
We use the remaining settings as those of CLIFF in Appendix \ref{sec:appendix:hyperparameters_and_dataset_stats}.
Note that we vary the sampling seed of the counterfactual samples in the multiple seed experiment.

To obtain the positive/negative summaries of the counterfactual document, we utilize the entities $E_o$ and $E_c$ used when obtaining the counterfactual document and apply the same entity replacement process to positive and negative summaries of the corresponding original document.
If there is no negative summary for the original document, we obtain it by performing entity replacement on $S_o$ with other counterfactual entities.
To gather a sufficient number of negative summaries, multiple counterfactual entity candidates are sampled during the process in Section \ref{subsubsec:factual_adaptiveness:evaluation_set_construction:counterfactual_entity_candidates} before the validation.

\subsection{Main Results}\label{subsec:experiments:results}
We compare the results of our model with those of NLL and CLIFF in Table \ref{table:ours_main_results} because CLIFF and ours sequentially apply additional techniques to NLL: contrastive learning and controllable counterfactual data augmentation, respectively.

From the perspective of conditional likelihood (i.e., $M_{CL}$), we can observe that our method significantly improves factual adaptiveness.
Compared to the contrastive learning baseline, our method also enhances factual consistency on the original test set in the BART-XSum case.

Although our models consistently reduce the $M_{CL}$ score, there is a case where our $M_{FC}$ score is higher than that of CLIFF in BART fine-tuned with CNN/DailyMail.
One possible explanation is that our method is more effective in terms of factual adaptiveness on datasets with a high level of abstractiveness such as XSum, while there is a misalignment between $M_{CL}$ and $M_{FC}$ on datasets with low abstractiveness (\citealp{
dreyer-etal-2023-evaluating}). We also provide the results of the ChatGPT preference test in Appendix \ref{sec:appendix:chatgpt_preference_test}.

\begin{figure}
    % \begin{minipage}{0.5\textwidth}
    \centering
    \includegraphics[width=0.45\textwidth]{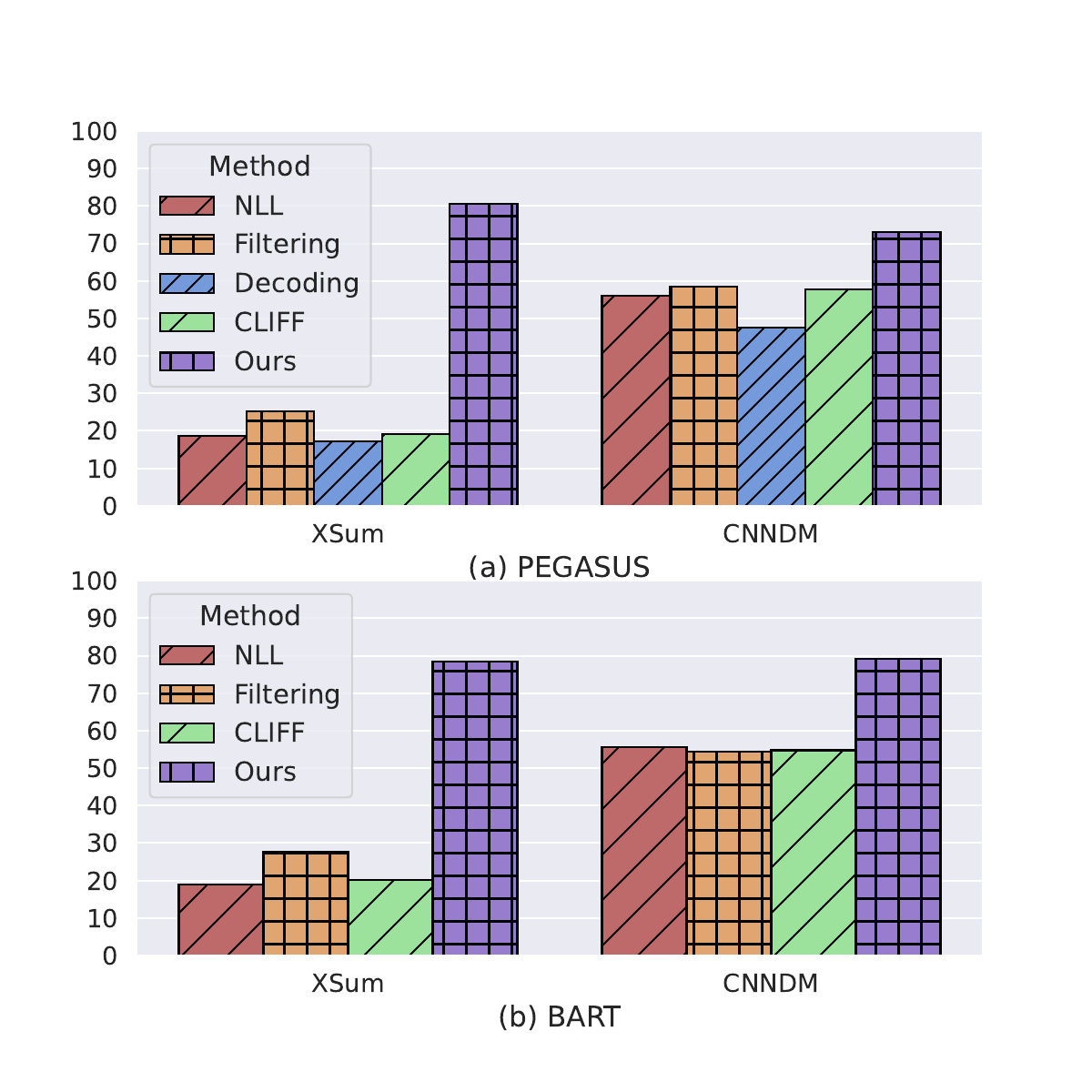}
    \caption{The ratio of summaries generated from the counterfactual documents of XSum and CNN/DailyMail (Mid, S2) which include the counterfactual entity but do not include the original entity.}
    \label{fig:entity_replacement_results}
    % \end{minipage}
\end{figure}

\section{Analysis}\label{sec:analysis}

\subsection{Entity Replacement}
The proportion of summaries that contain the counterfactual entity without the original entity given the counterfactual document is shown in Figure \ref{fig:entity_replacement_results}. 

We can observe that our model exhibits a significantly high rate of generating counterfactual entities in both datasets.
Filtering exhibits relatively higher values among the baselines, which is consistent with the results in Table \ref{table:baseline_analysis_pegasus}.
Compared to our method, however, Filtering still generates original entities at a high rate. 
The results also indicate that our approach successfully addresses entity-level hallucination problems in the BART-CNNDM setting where $M_{FC}$ is slightly higher than that of CLIFF.

\subsection{Counterfactual Entity Candidate Group}
We vary the counterfactual entity candidate group during the training set construction, as shown in Table \ref{table:M_CL_FC_ratio_group}.

It is observed that $M_{CL}$ scores are minimal when the group type of training set is aligned with the type of evaluation set.
We guess that the models tend to fit their factual adaptiveness to the distribution of training sets.
It is also observed that models fine-tuned with \textbf{Mid} group show low $M_{CL}$ scores across three evaluation sets.
Specifically, the score difference between the three evaluation groups of models fine-tuned with \textbf{Bot} group is the largest.
Based on those observations, we conclude that the distribution of counterfactual samples is important for entity-level generalization of factual adaptiveness.

\begin{table}
\centering
\resizebox{0.90\columnwidth}{!}{
\begin{tabular}{|c|c|c|c|c|c|c|}
\hline
\multicolumn{1}{|c|}{ } &
\multicolumn{1}{c|}{Aug.} &
\multicolumn{1}{c|}{Aug.} &
\multicolumn{3}{c|}{Evaluation Set} &
\multicolumn{1}{c|}{} \\\cline{4-6}
 & Group & Ratio & Top & Mid & Bot & QEval\\
\hline
\multirow{6}{*}{XSum} & \multirow{2}{*}{Top} & 5\% & 0.191 & 0.229 & 0.272 & 33.35 \\
& & 10\% & \textbf{0.147} & 0.188 & 0.236 & 33.37 \\\cline{2-7}
& \multirow{2}{*}{Mid} & 5\% & 0.217 & 0.162 & 0.156 & 33.34 \\
& & 10\% & 0.163 & \textbf{0.113} & 0.113 & \textbf{33.39} \\\cline{2-7}
& \multirow{2}{*}{Bot} & 5\% & 0.335 & 0.208 & 0.127 & 33.35 \\
& & 10\% & 0.285 & 0.159 & \textbf{0.081} & 33.37 \\
\hline
\multirow{6}{*}{CNNDM} & \multirow{2}{*}{Top} & 5\% & 0.142 & 0.153 & 0.152 & 51.04 \\
& & 10\% & \textbf{0.102} & 0.115 & 0.118 & 50.99 \\\cline{2-7}
& \multirow{2}{*}{Mid} & 5\% & 0.162 & 0.108 & 0.078 & 51.01 \\
& & 10\% & 0.118 & \textbf{0.073} & 0.047 & 51.01 \\\cline{2-7}
& \multirow{2}{*}{Bot} & 5\% & 0.205 & 0.124 & 0.060 & \textbf{51.08} \\
& & 10\% & 0.173 & 0.091 & \textbf{0.028} & 50.96 \\
\hline
\end{tabular}}
\caption{Mean of QuestEval (QEval) scores on original test sets and $M_{CL}$ scores on factual adaptiveness evaluation sets of our models based on BART varying the augmentation group (Aug. Group) and augmentation ratio (Aug. Ratio) across 3 seeds.
}
\label{table:M_CL_FC_ratio_group}
\end{table}

\begin{table}[!t]
\centering
\resizebox{0.91\columnwidth}{!}{
\begin{tabular}{l|ccccc}
\hline
\toprule
Dataset & NLL & Filtering & Decoding & CLIFF & Ours \\
\midrule
\multicolumn{6}{c}{PEGASUS} \\
\hline
XSum & \textbf{79.60} & 78.08 & 77.39 & 78.51 & 78.26\\
CNNDM & 11.44 & 9.67 & \textbf{13.61} & 11.23 & 11.21\\
\hline
\multicolumn{6}{c}{BART} \\
\hline
XSum & \textbf{80.17} & 78.76 & - & 79.45 & 79.32\\
CNNDM &\textbf{16.47} & 14.42 & - & 15.99 & 16.40\\
\hline
\end{tabular}}
\caption{
Mean of MINT scores across 3 seeds.
}\label{table:vs_abstractiveness}
\end{table}

\begin{table*}
\centering
\resizebox{\linewidth}{!}{
\begin{tabular}{|p{19cm}|}
\hline
Document (\textcolor{red}{Turkey}$\rightarrow$\textcolor{blue}{Portballintrae}) \\
\hline
Ece Heper, 50, was arrested on 30 December in the north-eastern town of Kars, her lawyer Sertac Celikkaleli told The Canadian Press. Canadian officials say they are offering consular assistance, but released no further information. ... \textcolor{blue}{Portballintrae}'s penal code states that anybody who insults the president can face up to four years in prison. … she was arrested for Facebook posts critical of President Recep Tayyip Erdogan. … \\
\hline
Summary \\
\hline
\textbf{NLL}: A Canadian woman has been charged with insulting the president of \textcolor{red}{Turkey}, her lawyer says. \\
\hline
\textbf{Filtering}: A Canadian woman has been charged with insulting the president of \textcolor{red}{Turkey}, her lawyer says. \\
\hline
\textbf{Decoding}: A Canadian woman has been arrested in \textcolor{red}{Turkey} for allegedly insulting the president of the \textcolor{blue}{Portballintrae} province, her lawyer says.\\
\hline
\textbf{CLIFF}: A Canadian woman has been arrested in \textcolor{red}{Turkey} on suspicion of insulting the president, her lawyer says. \\
\hline
\textbf{Ours}: A Canadian woman has been arrested in \textcolor{blue}{Portballintrae} on suspicion of insulting the president, her lawyer says. \\
\hline
\end{tabular}}
\caption{
Summaries of the counterfactual document of XSum (BART, Mid, S2) evaluation set.
Original and counterfactual entities are colored red and blue, respectively.
}\label{table:analysis:sample}
\end{table*}

\subsection{Augmentation Ratio}
We also vary the augmentation ratio $\rho$ which refers to the ratio of the size of the counterfactual samples to the size of the original training set in Table \ref{table:M_CL_FC_ratio_group}.
In all the cases, models of the augmentation ratio of 10\% exhibit much lower $M_{CL}$ scores compared to the augmentation ratio of 5\%, which implies that the degree of factual adaptiveness can be controlled by modifying $\rho$.
Interestingly, increasing $\rho$ does not always diminish the QEval scores while consistently enhancing factual adaptiveness.
The results reemphasize a close-to-orthogonal relationship between factual consistency and factual adaptiveness.

\subsection{Factual Adaptiveness vs. Abstractiveness}
To observe the relationship between factual adaptiveness and abstractiveness, we measure the MINT abstractiveness score (\citealp{dreyer-etal-2023-evaluating}) as shown in Table \ref{table:vs_abstractiveness}.
The abstractiveness of summaries generated by models fine-tuned with XSum demonstrates significantly higher levels of abstractiveness when compared to CNNDM, aligning with the findings of previous studies (\citealp{dreyer-etal-2023-evaluating}).

In the baselines, the lowest overall abstractiveness is found in Filtering with the highest factual adaptiveness.
On the other hand, our approach demonstrates a relatively minor trade-off between factual adaptiveness and abstractiveness.
The results suggest that our method substantially enhances factual adaptiveness while preserving the abstractiveness of generated summaries.

\subsection{Qualitative Study}
Table \ref{table:analysis:sample} shows summarization results given the counterfactual document where the entity \texttt{Turkey} is replaced by \texttt{Portballintrae}.
We use a model weight of BART\textsubscript{LARGE} provided by \textit{HuggingFace}\footnote[4]{\url{https://huggingface.co/facebook/bart-large-xsum}} to generate the sample for Decoding.
There are clues to infer \texttt{Turkey} such as \texttt{Kars} and \texttt{President Recep Tayyip Erdogan} which result in hallucinated summaries of baselines.
On the other hand, our model generates an accurate summary by adapting to the knowledge associated with \texttt{Portballintrae}.
We present another case study in Appendix \ref{sec:appendix:additional_samples}.
\section{Related Work}\label{sec:related_work}

\subsection{Factual Consistency of Summarization Models}
Studies on factual consistency of summarization models have been consistently conducted (\citealp{cao-wang-2021-cliff}; \citealp{wan-bansal-2022-factpegasus}; \citealp{rajagopal-2022-counterfactual}; \citealp{wan-etal-2023-faithfulness}; \citealp{roit-2023-factually}).
They enhance factual consistency through approaches from various directions such as post-editing (\citealp{chen-etal-2021-improving}; \citealp{balachandran-etal-2022-correcting}), data augmentation (\citealp{rajagopal-2022-counterfactual}), contrastive learning (\citealp{cao-wang-2021-cliff}; \citealp{wan-bansal-2022-factpegasus}), and advanced decoding (\citealp{king-2022-don}; \citealp{wan-etal-2023-faithfulness}).

\citet{rajagopal-2022-counterfactual} synthesize factually inconsistent summaries and augment the corresponding prompts to the document.
In this paper, we further modify input documents to trigger knowledge conflict effectively, analyze strategies to consider knowledge conflict, and demonstrate the robustness to entity-level knowledge conflict. 

There are also studies focusing on attributes other than factual consistency in summarization models (\citealp{west-etal-2022-probing}; \citealp{wu-etal-2022-frsum}; \citealp{cheang-2023-temposum}).
\citet{west-etal-2022-probing} analyze whether the model is grounded in the document by ablating facts related to the summary within the document.
\citet{wu-etal-2022-frsum} analyze the factual robustness, indicating whether the model assigns a low likelihood to an adversarial entity when given the document and factual prompt.

\subsection{Parametric Knowledge and Knowledge Conflict}
Recent studies in summarization have utilized general-purpose pre-trained language models (\citealp{lewis-etal-2020-bart}; \citealp{raffel-2020-exploring}; \citealp{brown-2020-language}; \citealp{ouyang-2022-training}; \citealp{chowdhery-2022-palm}; \citealp{chung-2022-scaling}) or have pre-trained the language model for summarization (\citealp{zhang-2020-pegasus}; \citealp{wan-bansal-2022-factpegasus}).

Recent studies have focused on addressing the hallucination problem in the language model caused by knowledge conflict, especially in question answering domain (\citealp{longpre-etal-2021-entity}; \citealp{neeman-2022-disentqa}; \citealp{li-2022-large}; \citealp{zhou-2023-context}).

\citet{ladhak-etal-2023-pre} and \citet{cheang-2023-temposum} analyze the hallucination problem of summarization models caused by knowledge conflict in a specific domain: name-nationality knowledge and evolving knowledge over time, respectively.
On the other hand, we analyze the robustness of summarization models concerning entity-level knowledge conflicts in arbitrary domains.
Moreover, we exploit parametric knowledge from PLM to effectively measure and improve factual adaptiveness.
\section{Conclusion}\label{sec:conclusion}
In this study, we analyze the factual adaptiveness of the fine-tuning based summarization models.
We propose two complementary metrics of factual adaptiveness and elucidate the relationship between factual consistency and factual adaptiveness.
We then propose a controllable counterfactual data augmentation method and observe that our method mitigates hallucination problems due to knowledge conflict.
Our experimental results show that our method effectively alleviates entity-level hallucination problems, especially when a knowledge conflict occurs.
We anticipate that our work will contribute to improving the faithfulness of summarization models that contain parametric knowledge.
\section*{Limitations}\label{sec:limitations}
In this paper, we conduct entity replacement to synthesize counterfactual samples to control knowledge conflict.
Because we utilize \textit{spaCy} to categorize named entity types, the performance of our method can vary depending on the accuracy of the tool.
We conduct research on PEGASUS and BART, and further investigation is needed regarding factual adaptiveness in large language models.
We focus on entity-level factual adaptiveness, and we leave expanding the scope of knowledge conflict as future work.
Future work can also consider orthogonal approaches such as decoding strategy, which can be integrated into our method.
\section*{Ethical Considerations}
We aim to improve the faithfulness of summarization models in terms of hallucination caused by knowledge conflict which is a major concern of (large) language model based approaches.
Our evaluation method could be used to diagnose parametric knowledge and factual adaptiveness which enhances the interpretability of the model.

\section*{Acknowledgments}
The authors are profoundly grateful for the support provided by the researchers at Samsung SDS. This work was supported by the National Research Foundation of
Korea (NRF) grant funded by the Korea government (MSIT)(2022R1A3B1077720 and 2022R1A5A708390811), Institute of Information \& Communications Technology Planning \& Evaluation (IITP) grant funded by the Korea government (MSIT) [2022-0-00959 and 2021-0-01343: Artificial Intelligence Graduate School Program (Seoul National University)], Samsung SDS Co.,Ltd, and the BK21 FOUR program of the Education and Research Program for Future ICT Pioneers, Seoul National University in 2024.

\bibliography{anthology,custom}

\clearpage
\appendix

\section{Transferability Test}\label{sec:appendix:transferability_test}
To clarify that the evaluation set construction method exploits parametric knowledge of PLM rather than global features such as word frequency, we additionally measure $M_{CL}$ and $M_{FC}$ on the evaluation set constructed from other PLM.
For example, we evaluate BART fine-tuned on XSum (i.e. BART (XSum)) with the evaluation set XSum (PEGASUS, Mid, S2).

The results of the transferability test are shown in Table \ref{table:transferability}.
$M_{CL}$ and $M_{FC}$ scores of misaligned cases in \textbf{Bot} group are lower than the aligned counterparts, implying that we also utilize parametric knowledge not only global attributes during the counterfactual sample synthesis.

\begin{algorithm}[t!]
\caption{Entity Validation Scenario (S2)}\label{alg:entity_selection_scenario_2}
\textbf{Input: } {Document $D_o=\{d_1, d_2, ..., d_M\}$, reference summary $S_o=\{s_1, s_2, ..., s_T\}$, pre-trained language model $\psi$, threshold $\tau$.} \\
\textbf{Output: } {Counterfactual samples $X_c$} \\
\begin{algorithmic}[1]
\newcommand*{\Break}{\textbf{break}}
\State $X_c = \{\}$
\State {Get $L=\{E_{o,1}, E_{o,2}, ..., E_{o,K}\}$, the list of named entity which exists in both $D_o$ and $S_o$}
\For{\texttt{$k \gets 1$ to $K$}}
    \State $t_k \gets $ {the first token position of $E_{o,k}$ in $S_o$}
    \State $E_c \gets $ {named entity sampled from one of three groups}\Comment{Section \ref{subsubsec:factual_adaptiveness:evaluation_set_construction:counterfactual_entity_candidates}}
    \State $D_c\gets$ \Call{Replace}{$D_o,E_{o,k},E_c$}
    \State $S_c\gets$ \Call{Replace}{$S_o,E_{o,k},E_c$}\Comment{Section \ref{subsubsec:factual_adaptiveness:factual_adaptiveness:entity_replacement}}
    \State $e_c \gets $ {the first token of $E_c$}
    \State $p_o \gets P_{\psi}(s_{t_k}|D_o, S_{o,<{t_k}})$
    \State $p_c \gets P_{\psi}(e_c|D_c, S_{c,<{t_k}})$
    \If {$p_o - p_c > \tau$}\Comment{Section \ref{subsubsec:factual_adaptiveness:factual_adaptiveness:evaluation_set_construction:original_and_counterfactual_named_entity}}
        \State {Append $(D_c, S_c)$ to $X_c$}
        % \State \Break
    \EndIf
\EndFor
\State \Return $X_c$
\end{algorithmic}
\end{algorithm}

\section{Algorithm of Entity Validation Scenario 2}\label{sec:appendix:algorithm2}
The detailed content of entity validation scenario 2 is presented in Algorithm \ref{alg:entity_selection_scenario_2}. 
The key difference with Algorithm \ref{alg:entity_selection_scenario_1} is that Algorithm \ref{alg:entity_selection_scenario_2} selects original and counterfactual entities, constructs counterfactual samples, and then calculates the conditional likelihood difference.

\section{Factual Adaptiveness of ChatGPT}
Factual adaptiveness evaluation results of ChatGPT are shown in Table \ref{table:chatgpt}.
We use \textit{gpt-3.5-turbo-0301} for ChatGPT and utilize PEGASUS to construct factual adaptiveness evaluation sets.

Because PLM which is used to construct factual adaptiveness evaluation sets is not aligned, there is no significant trend between the candidate groups in $M_{FC}$ due to the use of different PLM (i.e., PEGASUS) in the construction of factual adaptiveness evaluation sets.
We can observe that factual adaptiveness improves as the model size increases, but it is not completely resolved.

\begin{table}[!t]
\centering
\resizebox{\columnwidth}{!}{
\begin{tabular}{l|ccc|ccc}
\hline
\toprule
\multicolumn{1}{c}{} &
\multicolumn{3}{c}{$M_{CL}$(S2)($\downarrow$)} &
\multicolumn{3}{c}{$M_{FC}$(S2)($\downarrow$)} \\
Dataset & Top & Mid & Bot & Top & Mid & Bot \\
\midrule
\multicolumn{7}{c}{$\rightarrow$ BART (XSum)} \\
\hline
XSum (BART) & 0.762 & 0.752 & 0.757 & 2.09 & 2.14 & 2.26\\
XSum (PEGASUS) & 0.691 & 0.699 & 0.694 & 1.68 & 1.97 & 2.19\\
\hline
\multicolumn{7}{c}{$\rightarrow$ BART (CNNDM)} \\
\hline
CNNDM (BART) & 0.472 & 0.438 & 0.419 & 0.56 & 0.62 & 0.47\\
CNNDM (PEGASUS) & 0.380 & 0.357 & 0.327 & 0.65 & 0.59 & 0.47\\
\hline
\multicolumn{7}{c}{$\rightarrow$ PEGASUS (XSum)} \\
\hline
XSum (PEGASUS) & 0.718 & 0.734 & 0.744 & 2.14 & 2.43 & 2.55\\
XSum (BART) & 0.742 & 0.734 & 0.725 & 2.24 & 2.30 & 2.35\\
\hline
\multicolumn{7}{c}{$\rightarrow$ PEGASUS (CNNDM)} \\
\hline
CNNDM (PEGASUS) & 0.444 & 0.451 & 0.449 & 0.53 & 0.43 & 0.44\\
CNNDM (BART) & 0.458 & 0.424 & 0.401 & 0.59 & 0.50 & 0.37\\
\hline
\end{tabular}}
\caption{
Factual adaptiveness results (Scenario 2) when the fine-tuned PLM is aligned/misaligned with the model during the evaluation set construction.
}\label{table:transferability}
\end{table}

\begin{table}[!t]
\centering
\resizebox{\columnwidth}{!}{
\begin{tabular}{l|c|c|ccc|ccc}
\hline
\toprule
\multicolumn{1}{c}{} &
\multicolumn{1}{c}{} &
\multicolumn{1}{c}{} &
\multicolumn{3}{c}{$M_{FC}$(S1)($\downarrow$)} &
\multicolumn{3}{c}{$M_{FC}$(S2)($\downarrow$)} \\
Dataset & R-L & QEval & Top & Mid & Bot & Top & Mid & Bot \\
\midrule
\multicolumn{9}{c}{ChatGPT} \\
\hline
XSum & 20.74 & 43.54 & 1.68 & 1.53 & 1.55 & 1.38 & 1.48 & 1.51\\
CNNDM & 31.28 & 47.71 & 1.84 & 1.44 & 1.82 & 0.94 & 0.85 & 1.14\\
\hline
\multicolumn{9}{c}{Ours (CLIFF, PEGASUS)} \\
\hline
XSum & 35.69 & 33.26 & 1.35 & 1.23 & 1.32 & 1.29 & 1.20 & 1.38\\
CNNDM & 36.91 & 51.37 & 0.41 & 0.39 & 0.35 & 0.38 & 0.41 & 0.38\\
\hline
\end{tabular}}
\caption{
ROUGE-L (R-L) and QuestEval (QEval) scores on original test sets, and $M_{FC}$ scores of ChatGPT and ours on factual adaptiveness evaluation sets using PEGASUS. For ours, each score is the average value for 3 seeds.
}\label{table:chatgpt}
\end{table}

\begin{table*}[!t]
\centering
\resizebox{0.75\linewidth}{!}{
\begin{tabular}{l|c|c|ccc|ccc}
\hline
\toprule
\multicolumn{1}{c}{} &
\multicolumn{1}{c}{} &
\multicolumn{1}{c}{} &
\multicolumn{3}{c}{$M_{CL}$(S2)($\downarrow$)} &
\multicolumn{3}{c}{$M_{FC}$(S2)($\downarrow$)} \\
Models & R-L & QEval & Top & Mid & Bot & Top & Mid & Bot \\
\midrule
\multicolumn{9}{c}{XSum (BART)} \\
\hline
NLL & \textbf{34.83} & 32.94 & 0.762 & 0.752 & 0.757 & 2.09 & 2.14 & 2.26\\
& $\pm0.05$ & $\pm0.03$ & $\pm.004$ & $\pm.004$ & $\pm.002$ & $\pm0.14$ & $\pm0.02$ & $\pm0.14$\\
Filtering & 31.52 & \textbf{33.44} & \textbf{0.690} & \textbf{0.685} & \textbf{0.678} & \textbf{1.50} & \textbf{1.35} & \textbf{1.46}\\
& $\pm0.12$ & $\pm0.15$ & $\pm.005$ & $\pm.008$ & $\pm.015$ & $\pm0.04$ & $\pm0.16$ & $\pm0.08$\\
CLIFF & 33.89 & 33.32 & 0.748 & 0.742 & 0.747 & 2.18 & 2.19 & 2.36\\
& $\pm0.00$ & $\pm0.07$ & $\pm.002$ & $\pm.005$ & $\pm.004$ & $\pm0.19$ & $\pm0.07$ & $\pm0.17$\\
\hline
\multicolumn{9}{c}{CNN/DailyMail (BART)} \\
\hline
NLL & \textbf{38.05} & 50.99 & 0.472 & 0.438 & 0.419 & 0.56 & 0.62 & 0.47\\
& $\pm0.04$ & $\pm0.04$ & $\pm.005$ & $\pm.006$ & $\pm.005$ & $\pm0.10$ & $\pm0.10$ & $\pm0.11$\\
Filtering & 37.53 & \textbf{51.16} & \textbf{0.412} & \textbf{0.374} & \textbf{0.356} & 0.57 & \textbf{0.49} & \textbf{0.34}\\
& $\pm0.25$ & $\pm0.02$& $\pm0.008$ & $\pm0.011$ & $\pm0.012$& $\pm0.06$ & $\pm0.09$ & $\pm0.11$\\
CLIFF & 37.97 & 51.07 & 0.470 & 0.435 & 0.420 & \textbf{0.53} & 0.52 & 0.42\\
& $\pm0.13$ & $\pm0.06$ & $\pm.004$ & $\pm.003$ & $\pm.002$ & $\pm0.08$ & $\pm0.10$ & $\pm0.01$\\
\hline
\end{tabular}}
\caption{
ROUGE-L (R-L) and QuestEval (QEval) on original test sets and factual $M_{CL}$/$M_{FC}$ scores of BART on factual adaptiveness evaluation sets with the mean and standard deviation across 3 seeds.
}\label{table:baseline_analysis_bart}
\end{table*}

\begin{table*}[!t]
\centering
\resizebox{\linewidth}{!}{
\begin{tabular}{c|cc|cc|cc|cc}
\hline
\toprule
\multicolumn{1}{c|}{} &
\multicolumn{4}{c|}{PEGASUS} &
\multicolumn{4}{c}{BART}\\
\multicolumn{1}{c|}{} &
\multicolumn{2}{c}{XSum} &
\multicolumn{2}{c|}{CNN/DailyMail} &
\multicolumn{2}{c}{XSum} &
\multicolumn{2}{c}{CNN/DailyMail}\\
 & Scenario 1 & Scenario 2 & Scenario 1 & Scenario 2 & Scenario 1 & Scenario 2 & Scenario 1 & Scenario 2\\
\midrule
Threshold $\tau$ & 0.05 & 0.7 & 0.5 & 0.75 & - & 0.6 & - & 0.65 \\
\hline
\# Evaluation Set (Top) & 1,040 & 1,003 & 1,082 & 1,098 & - & 1,060 & - & 1,145 \\
\# Evaluation Set (Mid) & 1,041 & 1,163 & 1,079 & 1,411 & - & 1,339 & - & 1,659 \\
\# Evaluation Set (Bot) & 1,042 & 1,326 & 1,077 & 1,914 & - & 1,613 & - & 2,342 \\
\hline
\multicolumn{1}{c|}{\# Train Set (Original)} &
\multicolumn{2}{c|}{204,045} &
\multicolumn{2}{c|}{287,227} &
\multicolumn{2}{c|}{204,045} &
\multicolumn{2}{c}{287,227}\\
\multicolumn{1}{c|}{\# Train Set (Filtered)} &
\multicolumn{2}{c|}{74,241} &
\multicolumn{2}{c|}{159,519} &
\multicolumn{2}{c|}{74,241} &
\multicolumn{2}{c}{159,519}\\
\multicolumn{1}{c|}{\# Test Set (Original)} &
\multicolumn{2}{c|}{11,334} &
\multicolumn{2}{c|}{11,490} &
\multicolumn{2}{c|}{11,334} &
\multicolumn{2}{c}{11,490}\\
\hline
\multicolumn{1}{c|}{Learning Rate} &
\multicolumn{2}{c|}{1e-04} &
\multicolumn{2}{c|}{5e-05} &
\multicolumn{2}{c|}{3e-05} &
\multicolumn{2}{c}{3e-05}\\
\multicolumn{1}{c|}{\# Train Iter. (Filtered)} &
\multicolumn{2}{c|}{10k steps} &
\multicolumn{2}{c|}{110k steps} &
\multicolumn{2}{c|}{5 epochs} &
\multicolumn{2}{c}{5 epochs}\\
\multicolumn{1}{c|}{\# Train Iter. (Other)} &
\multicolumn{2}{c|}{30k steps} &
\multicolumn{2}{c|}{210k steps} &
\multicolumn{2}{c|}{5 epochs} &
\multicolumn{2}{c}{5 epochs}\\
\hline
\end{tabular}}
\caption{
Hyperparameters and data statistics.
}\label{table:hyperparameters}
\end{table*}

\section{Baseline Analysis on BART}\label{sec:appendix:baseline-analysis-bart}
Baseline analysis results on BART based models are shown in Table \ref{table:baseline_analysis_bart}.

We find that BART does not expose parametric knowledge in entity validation scenario 1.
Instead, we observe that replacing the \textit{null document} with the masked summary where named entities are replaced with a special [MASK] token reveals the parametric knowledge.
However, we do not further explore the optimal scenario for BART in this paper to provide general characteristics of fine-tuning based summarization models rather than model-specific analysis.
In addition, the tendency of increasing $M_{CL}$ scores in the order of \textbf{Top}, \textbf{Mid}, and \textbf{Bot} groups is observed to be low in BART.

\section{Hyperparameters and Dataset Statistics}\label{sec:appendix:hyperparameters_and_dataset_stats}
Threshold $\tau$ for each evaluation set and dataset statistics are shown in Table \ref{table:hyperparameters}.
Note that the size of the evaluation set of three groups is similar in Scenario 1 because we only use $E_o$ during the validation.

\textbf{Filtering} We exclude samples where at least one named entity in the summary does not appear in the document, except named entities of numerical categories.

\textbf{CLIFF} We choose \texttt{SysLowCon} setting used by \citet{cao-wang-2021-cliff}\footnote[5]{\url{https://github.com/ShuyangCao/cliff_summ}}. 
We use the same objective function and learning rates as those used in CLIFF except for the learning rate during the fine-tuning of PEGASUS with CNN/DailyMail; we use the initial learning rate of 5e-05 following \citet{zhang-2020-pegasus}.
We set the coefficient of contrastive loss to 1.0 and the batch size to 8 for both datasets.
Regarding the maximum number of negative samples, it is set to 5 for the XSum dataset and 4 for the CNN/DailyMail dataset.

\textbf{Advanced Decoding} We apply the method proposed by \citet{wan-etal-2023-faithfulness} to NLL models and follow \texttt{Beam + Greedy Lookahead} setup with a beam width 3\footnote[6]{\url{https://github.com/amazon-science/faithful-summarization-generation}}.
For the XSum dataset, we set the maximum output length to 60 and the look-ahead length to 16. 
For the CNN/DailyMail dataset, we set the maximum output length to 140 and the look-ahead length to 32.

\textbf{FactPEGASUS} We set the weight of contrastive loss to 5.0 and the maximum number of negative samples to 5.
We set the learning rate to 3e-05 and the training step to 15k following \citet{wan-bansal-2022-factpegasus}.
The batch size is set to 16, considering that the number of fine-tuning iterations in the original paper is half of that in CLIFF.

\section{Results on FactPEGASUS}\label{sec:appendix:factpegasus}
We follow hyperparameters in Appendix \ref{sec:appendix:hyperparameters_and_dataset_stats}.
Threshold $\tau$ is set to 0.35 for Scenario 2.

As shown in Table \ref{table:factpegasus_result}, our method can be effectively applied to FactPEGASUS as well.
Our method also slightly improves factual consistency on the original XSum dataset compared to the baseline.

\begin{table}[!t]
\centering
\resizebox{\columnwidth}{!}{
\begin{tabular}{c|c|c|cc}
\hline
\toprule
 & R-L & QEval & $M_{CL}$ & $M_{FC}$ \\
\midrule
FactPEGASUS & \textbf{27.06} & 34.02 & 0.597 & 1.59 \\
 & $\pm$0.04 & $\pm$0.09 & $\pm$.003 & $\pm$0.09 \\
 \hline
Ours & 26.79 & \textbf{34.12} & \textbf{0.149} & \textbf{1.16} \\
(FactPEGASUS) & $\pm$0.06 & $\pm$0.05 & $\pm$.004 & $\pm$0.04 \\
\hline
\end{tabular}}
\caption{
ROUGE-L (R-L) and QuestEval (QEval) on XSum test set and $M_{CL}$/$M_{FC}$ scores of on the factual adaptiveness evaluation set of Scenario 2 and \textbf{Mid} group with the mean and standard deviation across 3 seeds. 
}\label{table:factpegasus_result}
\end{table}

\section{ChatGPT Preference Test}\label{sec:appendix:chatgpt_preference_test}
Motivated by \citet{zhou-2023-lima}, we conduct a preference test using ChatGPT for the summaries generated by CLIFF and ours.
We use test sets of XSum and CNNDM for the experiment.
To remove ordering bias, we randomly shuffle the order of summaries of CLIFF and ours.

\begin{figure*}[t!]
\sbox\twosubbox{%
  \resizebox{\textwidth}{!}{%
    \includegraphics[height=5cm]{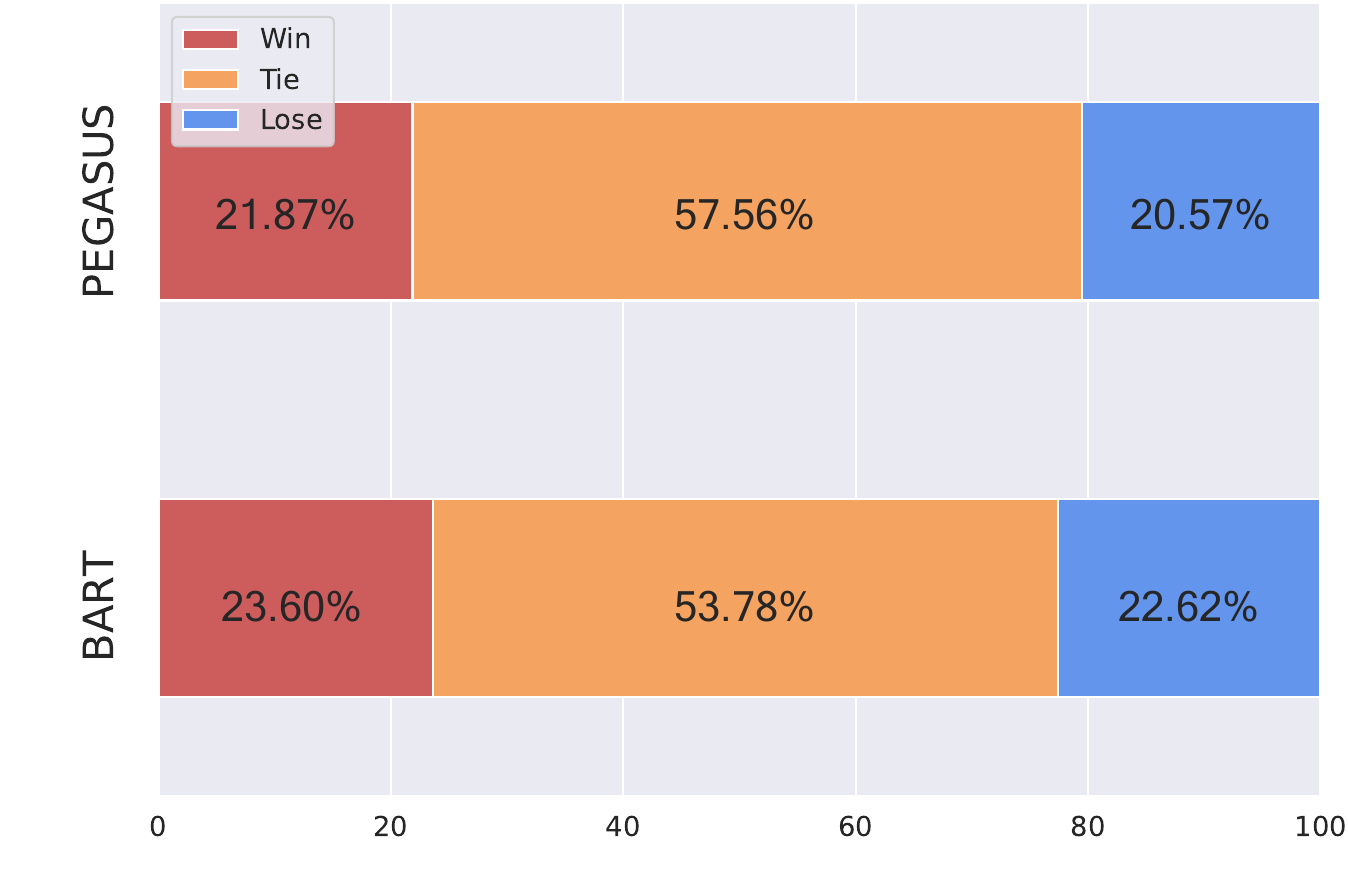}%
    \includegraphics[height=5cm]{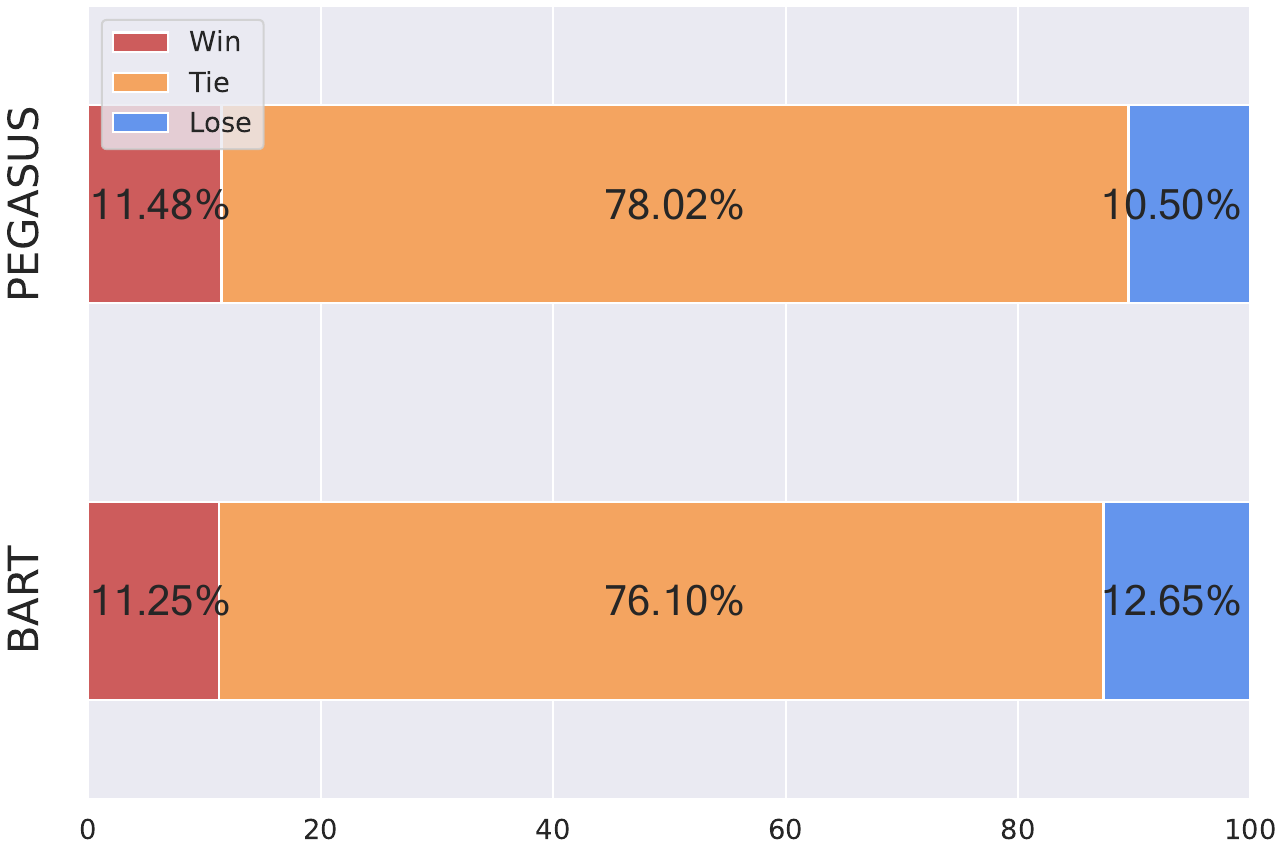}%
  }%
}
\setlength{\twosubht}{\ht\twosubbox}

\centering
\subcaptionbox{XSum\label{fig:3a}}{%
  \includegraphics[height=0.85\twosubht]{figures/figure3.1.pdf}%
}\hspace{15mm}
\subcaptionbox{CNN/DailyMail\label{fig:3b}}{%
  \includegraphics[height=0.85\twosubht]{figures/figure3.2.pdf}%
}
\caption{ChatGPT preference test results on (a) XSum and (b) CNN/DailyMail test sets.}\label{fig:chatgpt_preference_test}
\end{figure*}

The results are shown in Figure \ref{fig:chatgpt_preference_test}.
The term \textit{win} indicates that the summary generated by ours is preferred over that of CLIFF.
We observe a relatively high proportion of ties in the CNNDM.
We speculate that the results are attributed to the low abstractiveness of CNNDM, as mentioned in Section \ref{subsec:experiments:results}.
When compared to CLIFF, it is observed that ours generally generates preferred summaries for original documents.

\section{Additional Sample}\label{sec:appendix:additional_samples}
Other summarization examples are shown in Table \ref{table:appendix:additional_sample}.
Summaries of the baselines generate hallucinated entities instead of reflecting the counterfactual knowledge \texttt{Cherry Island}.
We speculate that the hallucinations are induced by the relevant entities such as the \texttt{UK} and \texttt{Northern Ireland}.

\section{License}
The repositories of \textit{fairseq}, FactPEGASUS, and XSum are under the MIT license.
The repositories of \textit{HuggingFace}, CLIFF, and CNN/DailyMail are under the Apache-2.0 license.
The repository of Decoding is under the CC-BY-NC-4.0, and MINT is under MIT-0.

\begin{table*}
\centering
\resizebox{\linewidth}{!}{
\begin{tabular}{|p{17cm}|}
\hline
Document (\textcolor{red}{London}$\rightarrow$\textcolor{blue}{Cherry Island}) \\
\hline
Lagmar Barking, a joint venture between MAR Properties and the Lagan Developments (Holdings), owned the Vicarage Field shopping centre in Barking. It has been bought by Benson Elliot, a UK-based private equity property fund manager. The last accounts for Lagmar Barking showed it owed its lenders £70m. The loan secured on the centre was bought by the US investment fund Cerberus as part of its purchase of the National Asset Management Agency's (Nama) Northern Ireland loan book in 2014. Peter Cornforth, director of retail at Benson Elliot, said the deal was "an exciting purchase for Benson Elliot". He added that it provided the firm with "a fantastic opportunity to contribute to the regeneration of a key east \textcolor{blue}{Cherry Island} metropolitan centre". Cerberus is continuing to rapidly work through the former Nama portfolio with a combination of asset sales, refinancings and enforcements.\\
\hline
Summary \\
\hline
\textbf{NLL}: A \hlcolor{whitered1}{Londonderry} shopping centre which was at the centre of a multi-million pound debt has been sold. \\
\hline
\textbf{Filtering}: A \hlcolor{whitered1}{Londonderry} shopping centre which went into administration last year has been sold for £\hlcolor{whitered1}{10m}. \\
\hline
\textbf{Decoding}:  The former owner of one of \hlcolor{whitered1}{Northern Ireland}'s largest shopping centres has been sold.\\
\hline
\textbf{CLIFF}: A shopping centre in \hlcolor{whitered1}{Londonderry} has been bought by a private equity firm. \\
\hline
\textbf{Ours}: A shopping centre in east \textcolor{blue}{Cherry Island} has been bought by a private equity firm for an undisclosed sum.\\
\hline
\end{tabular}}
\caption{
Summarization samples on counterfactual document based on XSum (BART, Mid, S2).
Hallucinated entities except for the original named entity $E_o$ are highlighted.
}\label{table:appendix:additional_sample}
\end{table*}

\end{document}